\documentclass[lettersize,journal]{IEEEtran}
\usepackage{amsmath,amsfonts}
\usepackage{algorithmic}
\usepackage{algorithm}
\usepackage{array}
\usepackage[caption=false,font=normalsize,labelfont=sf,textfont=sf]{subfig}
\usepackage{textcomp}
\usepackage{stfloats}
\usepackage{url}
\usepackage{verbatim}
\usepackage{graphicx}
\usepackage{cite}
\usepackage{wasysym} 
\begin{document}

\title{HydroelasticTouch: Simulation of Tactile Sensors with Hydroelastic Contact Surfaces}

\author{David P. Leins$^{1}$, Florian Patzelt$^{1}$ and Robert Haschke$^{1}$
\thanks{$^1$D. P. Leins, F. Patzelt and R. Haschke are with CITEC, Faculty of Technology, Bielefeld University, Germany. email: {\tt\small [dleins|fpatzelt]@techfak.uni-bielefeld.de}}%
}

\maketitle

\begin{abstract}
Thanks to recent advancements in the development of inexpensive, high-resolution tactile sensors, touch sensing has become popular in contact-rich robotic manipulation tasks.
With the surge of data-driven methods and their requirement for substantial datasets, several methods of simulating tactile sensors have emerged in the tactile research community to overcome real-world data collection limitations.
These simulation approaches can be split into two main categories: fast but inaccurate (soft) point-contact models and slow but accurate finite element modeling.
In this work, we present a novel approach to simulating pressure-based tactile sensors using the hydroelastic contact model, which provides a high degree of physical realism at a reasonable computational cost.
This model produces smooth contact forces for soft-to-soft and soft-to-rigid contacts along even non-convex contact surfaces. Pressure values are approximated at each point of the contact surface and can be integrated to calculate sensor outputs.
We validate our models' capacity to synthesize real-world tactile data by conducting zero-shot sim-to-real transfer of a model for object state estimation. Our simulation is available as a plug-in to our open-source, MuJoCo-based simulator.
\end{abstract}

\begin{IEEEkeywords}
Tactile sensor simulation, hydroelastic contact surfaces
\end{IEEEkeywords}

\section{Introduction}
Sensing and interpreting tactile feedback allows for identifying physical properties, including size, shape, texture, and dynamics, which are hard or impossible to perceive through other senses, e.g., vision.
As such, for both humans and robots, the sense of touch plays a pivotal role in successful, compliant interactions with their surroundings.

In the last decade, various tactile sensors with different sensing modalities have been developed, and the field of robotics has seen a surge in tactile-enabled robotic applications.
See \cite{mandilTactileSensingTechnologiesTrends2023} for a comprehensive review of tactile sensing modalities and their applications.

Primarily, vision-based tactile sensors, generally consisting of an elastic membrane and one or more cameras capturing the deformation of the membrane, have been very prominent in recent research due to their cost-effective design and high-resolution output \cite{wangHSVTacHighSpeedVisionBased2023,piacenzaSensorizedMulticurvedRobot2020,gomesGelTipFingershapedOptical2020,lambetaDIGITNovelDesign2020,padmanabhaOmniTactMultiDirectionalHighResolution2020}. Nevertheless, common drawbacks of optical sensors are their bulkiness, lower sampling frequencies, and lower first-touch sensitivities. Thus, resistive sensors remain a relevant alternative when high frequency, sensitivity, and compact and flexible design have a higher priority. In the latter category, piezoresistive sensors have become the most common strategy for tactile perception due to their simple structure, flexible manufacturing, straightforward read-out mechanisms, and high sensitivity \cite{zhuRecentAdvancesResistive2022}.

\begin{figure}[t]
    \centering
    \includegraphics[width=\linewidth]{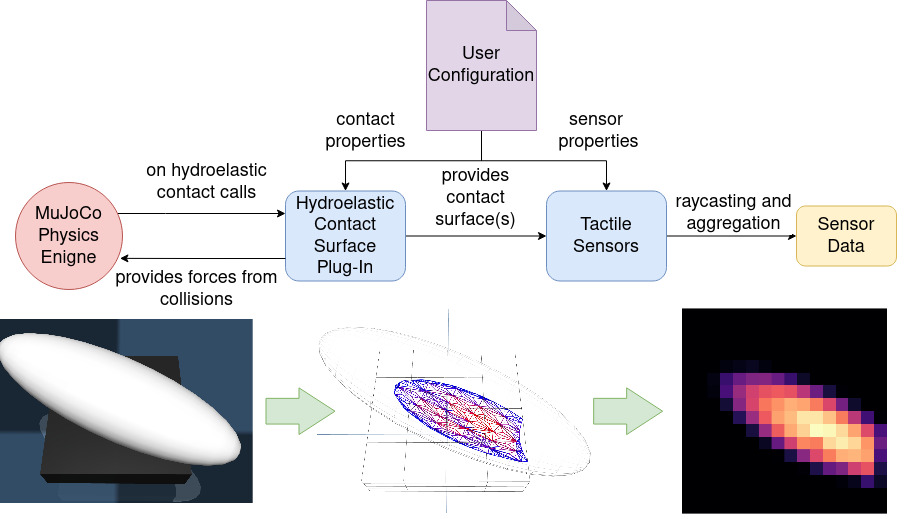}
    \caption{Information flow of the proposed simulation pipeline. The MuJoCo engine calls our plug-in on a detected hydroelastic contact, which computes the contact dynamics based on the user configuration. This step yields a) forces the engine uses to advance the simulation and b) provides a sensor involved in a collision with the respective contact surface. Each sensor instance then samples pressures from the surface and derives its readings based on the user configuration.}
    \label{fig:overview}
\end{figure}

\IEEEpubidadjcol
With the growing popularity of machine learning-based sensor processing methods, running real-world experiments to gather training data has become increasingly infeasible.
Simulation provides a controlled, cost-effective, and safe environment for researchers to explore, analyze, and collect data without needing physical prototypes or manual setup routines and with no risk of wear and tear of hardware components.
As such, simulation has become an increasingly important tool in response to the extensive dataset requirements of data-driven methods.
Sim-to-real transfer has become widely adopted, especially for sample inefficient methods such as Deep Reinforcement Learning (DRL).
We refer to \cite{zhaoSimtoRealTransferDeep2020} for a recent review.

In recent years, the tactile research community has seen a variety of new approaches to simulating vision-based sensors \cite{gomesGelSightSimulationSim2Real2019,dingSimtoRealTransferOptical2020,sferrazzaLearningSenseTouch2020,churchTactileSimtoRealPolicy2022,siTaximExampleBasedSimulation2022,wangTACTOFastFlexible2022,xuEfficientTactileSimulation2022,chenTacchiPluggableLow2023,chenGeneralPurposeSim2RealProtocol2024} and a few to simulate non-vision-based sensors \cite{narangInterpretingPredictingTactile2020,narangSimtoRealRoboticTactile2021,kappassovSimulationTactileSensing2020}.
For tactile sensor simulation, Wang et al. \cite{wangTACTOFastFlexible2022} propose high throughput, flexibility, realism, and ease of use as desiderata.
From the perspective of robotic learning research, we especially emphasize high throughput and realism in the examining of existing work.
With this consideration, we notice that while Finite Element Method (FEM) based approaches are highly accurate, they are infeasible for real-time simulation and data-hungry learning methods due to their heavy computational effort.
On the other hand, the remaining methods use oversimplified physical models that greatly increase the simulation speed but come at the cost of unrealistic data and require intricate task and sensor-specific configuration, data augmentation, domain randomization, or even specific translation models for successful transfer of learned models into the real-world domain.\\

This work presents a novel method capable of simulating arbitrarily shaped sensors with customizable material parameters. By integrating the hydroelastic contact model \cite{elandtPressureFieldModel2019} into the simulation's physics update, we compute more realistic contact forces for soft-to-rigid and soft-to-soft body collisions than point contact models. The hydroelastic contact model calculates contact surfaces and pressure values on this surface. We aggregate samples of these pressure values with a fully parameterizable model to derive sensor cell values. The computational effort, linked to the simulation's degree of realism, can be scaled by several parameters, e.g., a resolution parameter to adapt the simulation model to the specific object, sensor, or task requirements.
Fig.~\ref{fig:overview} illustrates the information flow of our proposed method.

In an experimental evaluation, we train a neural network to predict object angles based on the sensor response of a simulated piezoresistive sensor and validate our method's ability to simulate such sensors by successfully performing zero-shot sim-to-real transfer.

Our tactile sensor simulation is available as a plugin\footnote{\url{https://github.com/ubi-agni/mujoco_contact_surfaces}} to our open source simulator using the MuJoCo physics engine \cite{todorov2012mujoco}.

\section{Related Work}
In this section, we discuss present work in tactile sensor simulation, the differences between approaches, and their advantages and disadvantages.

Narang et al. use a soft material model of the BioTac sensor \cite{wettelsBiomimeticTactileSensor2008} and apply FEM for realistic simulation of high-density touch in combination with data-driven learning \cite{narangInterpretingPredictingTactile2020}, but even with the later GPU-acceleration in \cite{narangSimtoRealRoboticTactile2021} simulation time is still not feasible for real-time robotic simulation (up to 5.6s for a simulation step). Another shortcoming is that their model is hand crafted for a BioTac sensor, and because of manufacturing differences, even for BioTac sensors, instance-wise fine-tuning is necessary.
Sferrazza et al. take a similar approach of training a deep neural network with FEM-generated data to predict the surface deformation of a marker-based optical tactile sensor \cite{sferrazzaLearningSenseTouch2020}. However, this involves recording a sensor-specific dataset, which the trained model is limited to. 

In \cite{gomesGelSightSimulationSim2Real2019}, Gomes et al. compute the sensor image of an optical tactile sensor by applying Phong's reflection model to the reconstructed elastomere surface, approximated by a simulated depth camera capturing the extent to which objects penetrate the elastomere. This approach is limited to touching fixed objects and does not consider any contact dynamics.
Ding et al.~\cite{dingSimtoRealTransferOptical2020} apply an elastic deformation model directly on the simulated tip of a TacTip sensor \cite{ward-cherrierTacTipFamilySoft2018} through the Unity physics engine. They report that the Unity engine restricts the simulation speed.
With an internal deformation function, Wang et al.~\cite{wangTACTOFastFlexible2022} map the contact forces of any underlying physics engine to the gel deformation of an optical tactile sensor and provide a configuration framework to simulate a set of different optical tactile sensors. However, this approach highly depends on the underlying engine's contact dynamics and only targets vision-based sensors.
Xu et al.~\cite{xuEfficientTactileSimulation2022} resort to a penalty-based rigid body dynamics model to simulate an optical tactile sensor's differential normal and shear force. Yet, this approach struggles with correctly approximating soft sensors.
Kappassov et al.~\cite{kappassovSimulationTactileSensing2020} propose a simulation framework for tactile sensing arrays by interpreting rigid-body point contacts as point stimuli and computing the array's response heuristically with a point spread function.

A model for an elastic interaction of particles has been suggested by Wang et al.~\cite{wangElasticTactileSimulation2021} to simulate the elasticity of sensors and touched objects. They compute the deformation by converting the meshes to particle clouds with elastic properties. The deformation is then used to generate a depth image.
Church et al.~\cite{churchTactileSimtoRealPolicy2022} suggest intentionally using only depth image rendering for swift data generation and later training a domain adaption neural network to map simulation data to the real world. However, while the simulation data can be generated relatively fast, this approach's bottleneck lies in gathering matching (possibly task-specific) data in both domains and the requirement of training the translation model.

As limitations in the current state of the art, we identify that data generation is too slow for methods with a higher degree of realism. In contrast, faster methods rely on inaccurate rigid-body assumptions or slightly more accurate soft-body point contact models. These inaccuracies cause a higher sim-to-real gap that must be narrowed with various regularization methods, manifesting in longer training times. Additionally, most methods are targeted at optical tactile sensors, which can not trivially be adapted to yield physical quantities directly, such as the pressure distribution over a sensor surface.

The novelty in our approach lies in replacing (soft body) point contacts with physically more accurate hydroelastic contact surfaces, trading off a configurable amount of computational effort with higher simulation accuracy. While our approach is currently targeted at non-optical tactile sensors, it can be extended to optical tactile sensors by rendering the computed deformation of the sensor surface.

\section{Methods}

This section describes the integration of the hydroelastic contact surface model into our simulation framework and how sensor data can be derived.

\subsection{Hydroelastic Contact Surfaces}
As introduced in~\cite{elandtPressureFieldModel2019},
the hydroelastic contact model integrates two fundamental concepts: elastic foundation and hydrostatic pressure. To model hydrostatic pressure, a virtual pressure field $p_O(x)$ must be pre-computed for each object, wherein the pressure is maximal at the center, diminishes to zero at the boundary, and is interpolated in the intervening regions.

The computation of the pressure field for arbitrary objects involves solving Laplace's equation over a Cartesian domain, subject to boundary conditions dependent upon the object's shape; additional details can be found in \cite{elandtPressureFieldModel2019}. Each object's pressure field is approximated using a coarse tetrahedral volume mesh, storing pressure values at the vertices and employing interpolation in the intermediate areas.

In the simulation process, a contact surface is computed when two bodies collide. For two overlapping objects, A and B, with pressure fields $p_A(x)$ and $p_B(x)$, the contact surface $S$ is defined as the surface of equal pressure: $p_e(x) = p_A(x) = p_B(x)$. This contact surface is represented by a triangle mesh, where pressure values are stored at each vertex and can be interpolated between them.

Forces and moments acting on colliding bodies are computed by integrating pressure values over the contact surface. This is achieved by approximating the force contribution from each triangle. The integral over the linear pressure field gives the elastic force contribution from a triangle with area $A$:

$$f_e = \int_A p_e(x) \hat{n} dA = A p_e(x_c) \hat{n} $$

Here $\hat{n}$ is the normal direction, and $p_e(x_c)$ is the pressure evaluated at the centroid $x_c$ of the triangle. To impart an energy-damping property, Hunt-Crossley-dissipation at the stress level $d$ is included in the model: 

$$f_e = A  p_e(x_c) (1 - v_n)  \hat{n} $$

Here $v_n$ is the contact velocity in the normal direction.
The pressure is approximated using a Taylor expansion in time to compute contact forces and the resulting dynamics. For more details, please refer to~\cite{masterjohnVelocityLevelApproximation2022}.

In our implementation hydroelastic contact for an object is enabled by specifying five parameters: The \textit{hydrostatic modulus}, which defines the maximum value of the object's pressure field $p_O(x)$ in $N/m^2$, consequently determining its stiffness, the \textit{Hunt-Crossley dissipation} (measured in $s/m$) and \textit{coefficients for static and dynamic friction}. In addition, for most primitive object shapes, a \textit{resolution hint} is required. These objects must be tessellated into meshes to compute their pressure field. The resolution hint controls the fineness of the meshes.

\begin{figure}[tbp]
    \centering
    \subfloat[]{\includegraphics[width=0.45\linewidth]{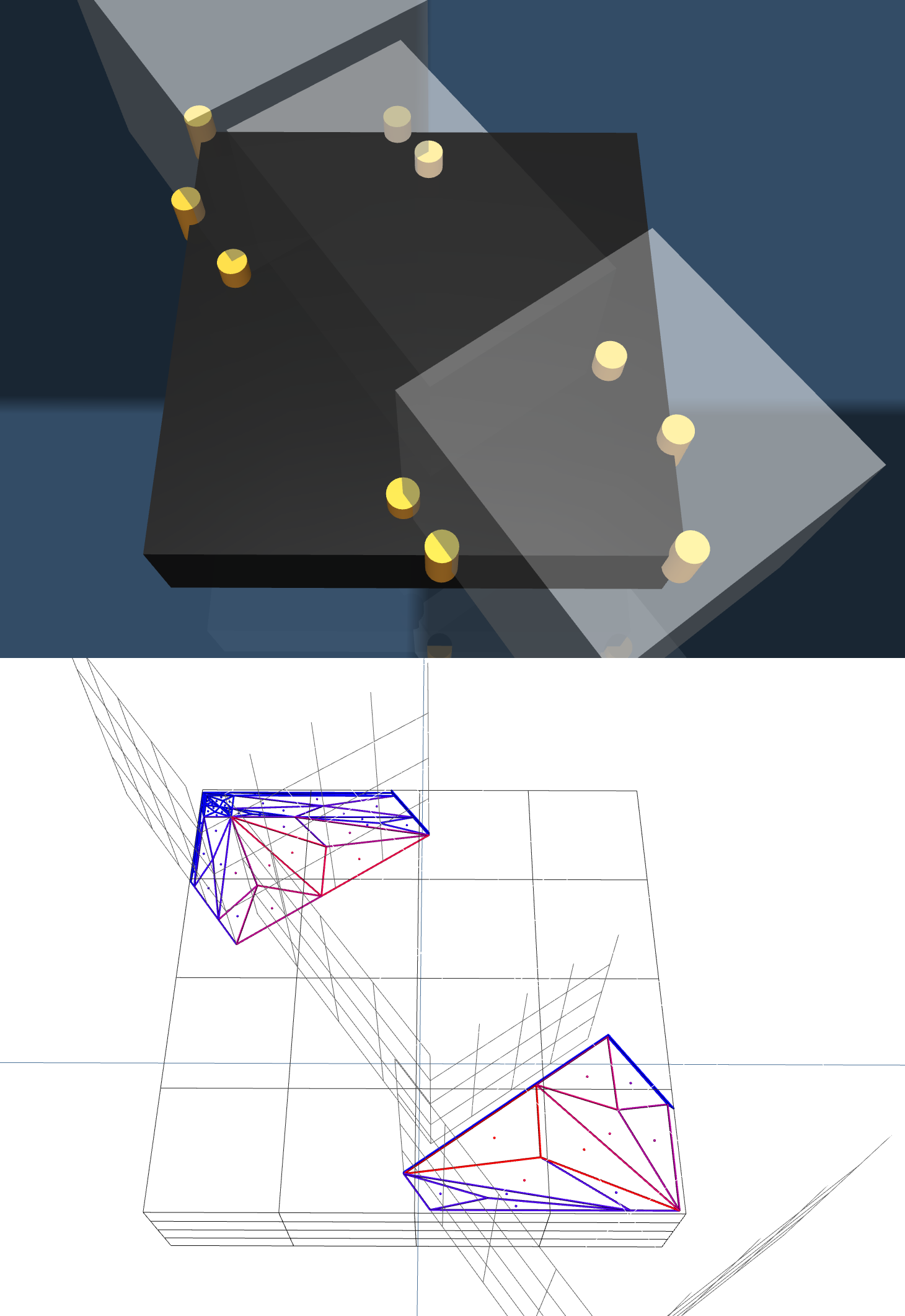}}
    \subfloat[]{\includegraphics[width=0.45\linewidth]{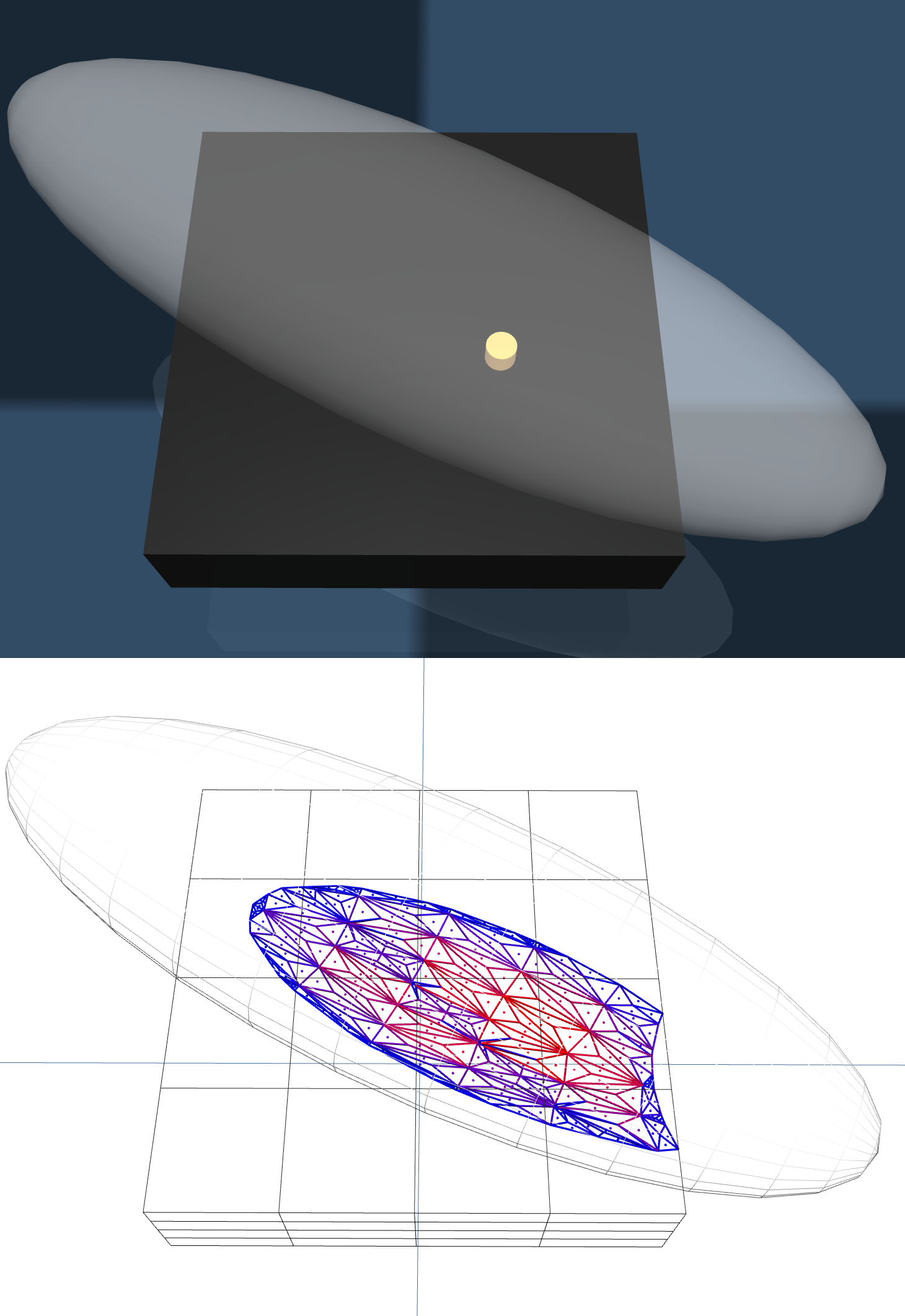}}
    \caption{Comparison between default MuJoCo point contacts and contact surfaces with (a) a cuboid geometry with a convex contact surface and (b) an ellipsoid geometry with a concave contact surface. Yellow cylinders depict point contact locations. In the lower part, the contact surface meshes are depicted (in wireframe mode for better visibility) and colored from blue (low pressure) to red (high pressure) based on the normalized pressure range experienced in the respective collision.}
    \label{fig:pc-vs-cs}
\end{figure}

Our simulator parses the model and automatically overrides the contact physics with a drop-in replacement when two bodies with hydroelastic parametrization collide.
Note that the faster default point contact model computes the collision dynamics for any contact pair involving only one or no parameterized bodies.
The difference between the engine's default point contact and the contact surface for an exemplary collision is illustrated in figure~\ref{fig:pc-vs-cs}.

\subsection{Sensor Definition}
A simulated sensor is defined by a body, i.e., a primitive geometry or a mesh, for which hydroelastic collisions are enabled.
Additionally, a sensor must define a set of points in 3D space (on or close to the geometry's surface) and a set of normal vectors corresponding to the locations of sensor cells and their respective primary axis of pressure sensitivity.
Further, a sensor allows configuring a sampling resolution (measured in $1/m^2$), an aggregation function $\phi$ to apply on the sampled values, and the receptive radius of a cell $r_c$. We opt for a receptive region around the cell given by the receptive radius instead of specifying physical cell extents to reflect that forces are usually distributed isometrically on a sensor, regardless of cell shape.
A sensor works by sampling pressure values from a contact surface and then aggregating them to calculate the sensor readings. We first introduce the sampling process and then describe the sensor value calculation.\\

\subsection{Deriving Pressures from the Contact Surface}
\label{sec:deriving_pressure}
The contact surface computed for a hydroelastic collision is represented as a triangle mesh, which provides a piece-wise linear pressure field on its triangles. These values must then be processed and mapped to the coordinates on or near the sensor mesh surface defined by the cell positions.
We solve the problem of projecting the contact surface's pressure values to these cell positions with raycasting, where the closest intersection of the surface's triangle mesh and a ray originating at a cell center, cast into the direction of its normal, is computed.
For this kind of problem, we can apply well-researched and optimized ray-triangle intersection algorithms \cite{moller2005fast} that can be parallelized cell-wise and be further sped up with bounding volume hierarchies \cite{wald2007fast}.
When a valid intersection with the contact surface mesh is found, the value of the pressure field at the intersection point can be directly accessed. If no valid intersection can be found, no pressure is measured.

However, a single sample might not be accurate enough to approximate the pressure exerted on a cell. Thus, we opt for a sampling-based method to approximate the integral of the pressure values that should lie in a cell's receptive sphere.
The sample resolution parameter mentioned earlier gives the sample size, which we distribute uniformly on the mesh surface. In the trivial case of a plane, this corresponds to a regular two-dimensional grid. In contrast, for non-trivial meshes, we achieve a uniform distribution on the surface with Constrained Poisson-disk sampling \cite{sampling}. Each sensor cell then considers the subset of sampled coordinates in its receptive sphere. These coordinates can be constant and thus are computed in a pre-processing step during the initialization of the simulation. Note that a single sample point can be attributed to different sensors.

Optionally, two functions can be defined to weigh the contribution of the sample based on the distance to the cell center and the angle between the sensor mesh's normal at the sample coordinate and the cell's normal. We found the squared geodesic distance for the former and, for the latter, a filter considering only samples within $[-\pi/4,~\pi/4]$ to be reasonable defaults.\\

Finally, with the given definitions, we can formalize computing the value $v_c$ of a sensor cell $\mathcal{C}$ with algorithm \ref{alg:cell}.

\begin{algorithm}[b]
\caption{Computation of pressure value $v_c$ for a sensor cell $\mathcal{C}$ with sampled rays $\mathcal{R}$, contact surface $\mathcal{S}$, combined spatial and angular distance weight vector $\Vec{\beta}$, and aggregation function $\phi$.}
\begin{algorithmic}
    \FOR{$r_i \in \mathcal{R}$}
    \STATE $q \gets$ find\_intersection$(r_i, \mathcal{S})$
    \IF{is\_finite$(q)$}
    \STATE $p_i \gets \beta_i \cdot$ surface\_pressure\_at$(q, \mathcal{S})$
    \ELSE
    \STATE $p_i \gets 0$
    \ENDIF
    \ENDFOR
    \STATE $v_c \gets \phi(\Vec{p})$
    \end{algorithmic}
    \label{alg:cell}
\end{algorithm}

\begin{figure}[tbp]
    \centering
    \subfloat[]{\scalebox{-1}[1]{\includegraphics[width=0.45\linewidth]{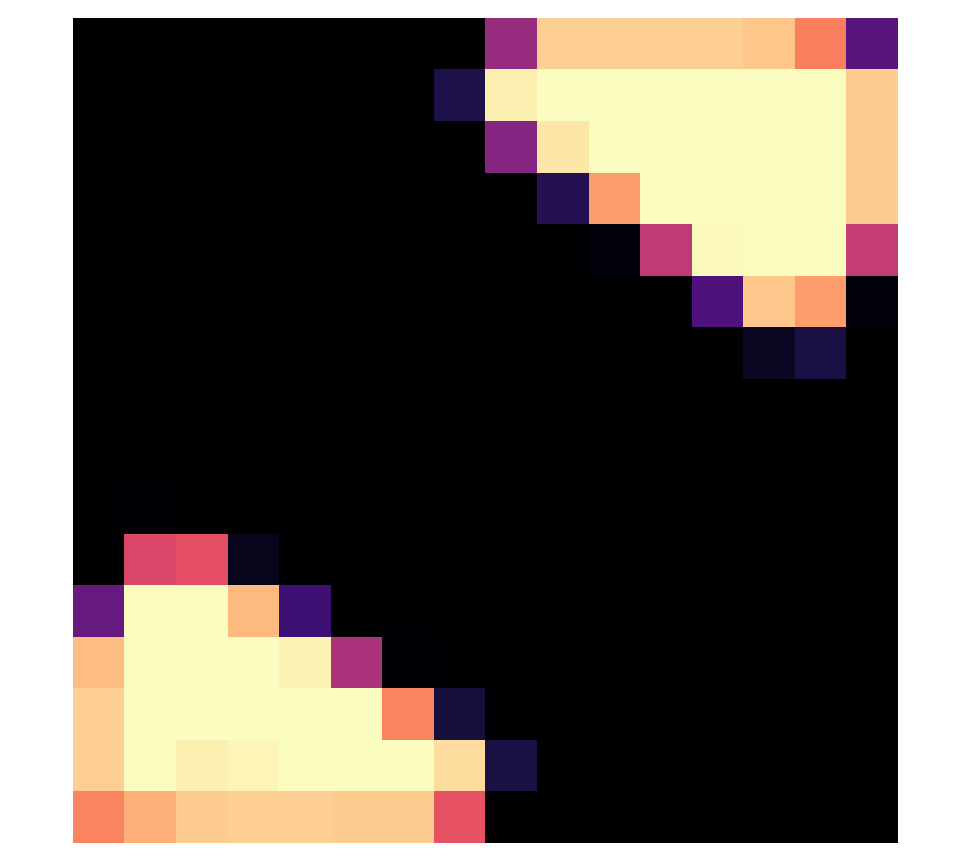}}}
    \subfloat[]{\scalebox{-1}[1]{\includegraphics[width=0.45\linewidth]{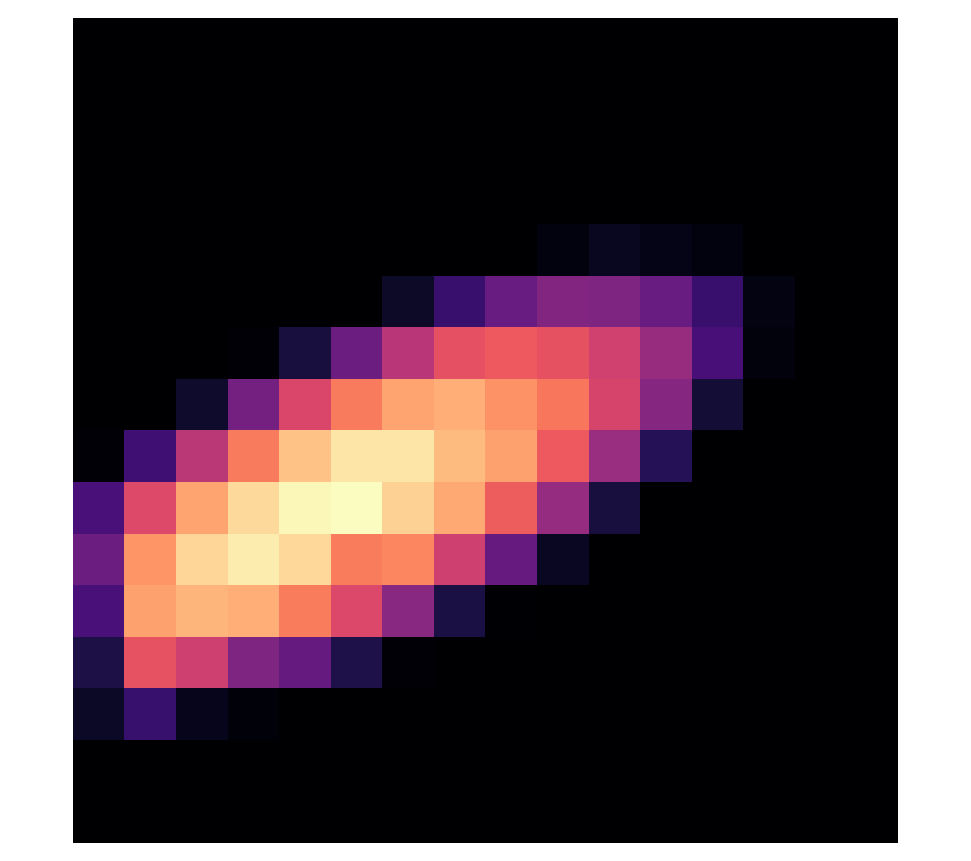}}}
    \caption{Exemplary sensor images derived from the contact surfaces in figure~\ref{fig:pc-vs-cs}.}
    \label{fig:pc-sensor}
\end{figure}

Figure~\ref{fig:pc-sensor} depicts the resulting sensor readings derived by our method for the object placements visualized in figure~\ref{fig:pc-vs-cs}.

\subsection{Trade-off between Realism and Simulation Speed}
The sensor configuration allows flexible fine-tuning of the simulation behavior according to the user's requirements. These primarily impact simulation accuracy and speed, balancing one against the other.
We provide a brief overview of these customization options in the subsequent sections. 

\subsubsection{Sensor Frequency}
The raycasting process requires many computations, and although they are run in parallel and are optimized in vectorized operations, the amount of CPU threads can not keep up with the number of cast rays for larger sensors or higher resolutions.
Reducing the sensor frequency, i.e., the rate at which the contact surface pressures are mapped to sensor data, smooths out the computational load. For tasks where sensor data is only required on demand, we also offer a mode where sensor data is only computed when explicitly requested, further maximizing simulation speed.
Contact surfaces are still computed for every simulation step to ensure consistent physics regardless of sensor frequency.

\subsubsection{Mesh Resolution}
The resolution of meshes used to compute the static pressure fields of simulated bodies can drastically influence the speed and accuracy of the simulation. For primitive objects, this resolution can be defined with the resolution hint parameter. While a finer resolution of the pressure field can enhance contact surface resolution, facilitating more precise sensor readings upon sampling, it simultaneously raises the computational demand. Specifically, increasing the number of elements within the tetrahedral volume mesh representing the pressure field also increases the number of intersection tests computed to compile a contact surface from two colliding volume meshes.

\subsubsection{Sample Resolution}
Depending on the tessellation and resolution of the contact surfaces, there might be smaller triangles in some areas of the contact surface.
The sample resolution commands the amount of sampled rays used for raycasting. A sample resolution that is too coarse can miss these smaller triangles, which then do not contribute to the sensor value. A more thorough analysis of pressure distribution over various contact surfaces uncovered that the algorithm computing the pressure field sometimes creates uneven pressure distributions in a local triangle neighborhood. While the integral over these neighborhoods yields the expected pressure values, very small triangles tend to get a higher pressure than their bigger neighboring triangles. We found these abnormal regions prominently at the edges of geometries and along a primary tetrahedral decomposition of its pressure volume, e.g., one of two diagonals of a cuboid. This behavior explains the anomalies we noticed in a preliminary study \cite{leins2023more}, where we measured slightly lower pressures along the diagonal of a sensor.
Increasing the sampling resolution will reduce inaccuracies caused by this behavior but also result in a heavier computational load.

\section{Experimental Setup}

\begin{figure}
    \centering
    \includegraphics[height=0.5\linewidth]{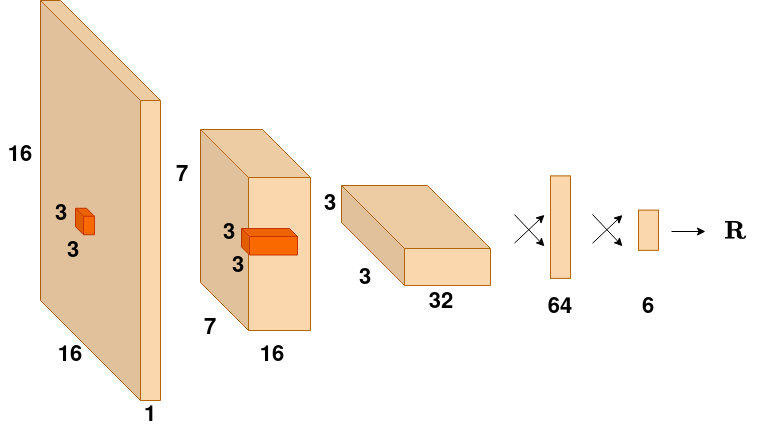}
    \caption{The used architecture to predict a rotation matrix $\textbf{R}$, which in our case corresponds to $\textbf{R}^W_{O'}$. Both convolutional layers use a stride of 2. The crossed arrows depict fully connected layers. The output is interpreted as a 6D representation of a rotation matrix comprising the first two columns.}
    \label{fig:nn-model}
\end{figure}

We validate the quality of our simulation method by performing zero-shot sim-to-real transfer. Specifically, we train a neural network to estimate object states using synthetic tactile sensor data. Then we assess the model's performance on real-world data from objects not encountered during training. The experimental setup is inspired by Lach et al. \cite{lachPlacingTouchingEmpirical2023}, where two piezoresistive sensors \cite{SchurmannMyrmex} are attached to the inside of a two-finger gripper. Lach et al. use a neural network to predict the orientation of an object in the gripper with the tactile data to place it steadily on a planar surface.
We modify the experimental configuration to employ a single sensor of the same type. Data is collected using a Franka Emika Panda robot that presses objects onto the sensor, enabling us to deduce the object rotation from the gripper frame orientation directly.

Following the nomenclature of Lach et al., the object's placing normal $\Vec{z}_p = \textbf{R}^W_{O'} \Vec{z}$ is defined as the z-axis in the local object frame $O'$, where the rotation from the world reference frame to $O'$ is defined as $\textbf{R}^W_{O'} \in SO(3)$. As in our adaptation, the sensor is fixed, and we choose the sensor frame as the reference frame. Further, we derive the $O'$ frame by rotating the panda's gripper frame with a constant rotation $\textbf{R}^G_{O'} \in SO(3)$.

In our experiment, we use the same model architecture used in \cite{lachPlacingTouchingEmpirical2023} (see figure~\ref{fig:nn-model}) to predict the rotation $\textbf{R}^W_{O'}$ with the following two adaptations: First, we change the input channel dimension of the first layer to 1, as we only use a single sensor input. Second, we adapt the original loss function to account for the symmetry in the computation of the angle difference

$$\mathcal{L}(\textbf{R}, \Vec{z}^{gt}_p) = \min (|cos^{-1}(\textbf{R}\Vec{z}\cdot\Vec{z}^{gt}_p)|, |cos^{-1}(\textbf{R}\Vec{z}\cdot(\Vec{z}^{gt}_p + \pi)|)$$

with \textbf{R} being the predicted rotation matrix, and $\Vec{z}^{gt}_p$ the measured ground truth placing.

\subsection{Real-World Dataset}
\label{sec:real-data}
\begin{figure}[tp]
    \centering
    \includegraphics[height=0.5\linewidth]{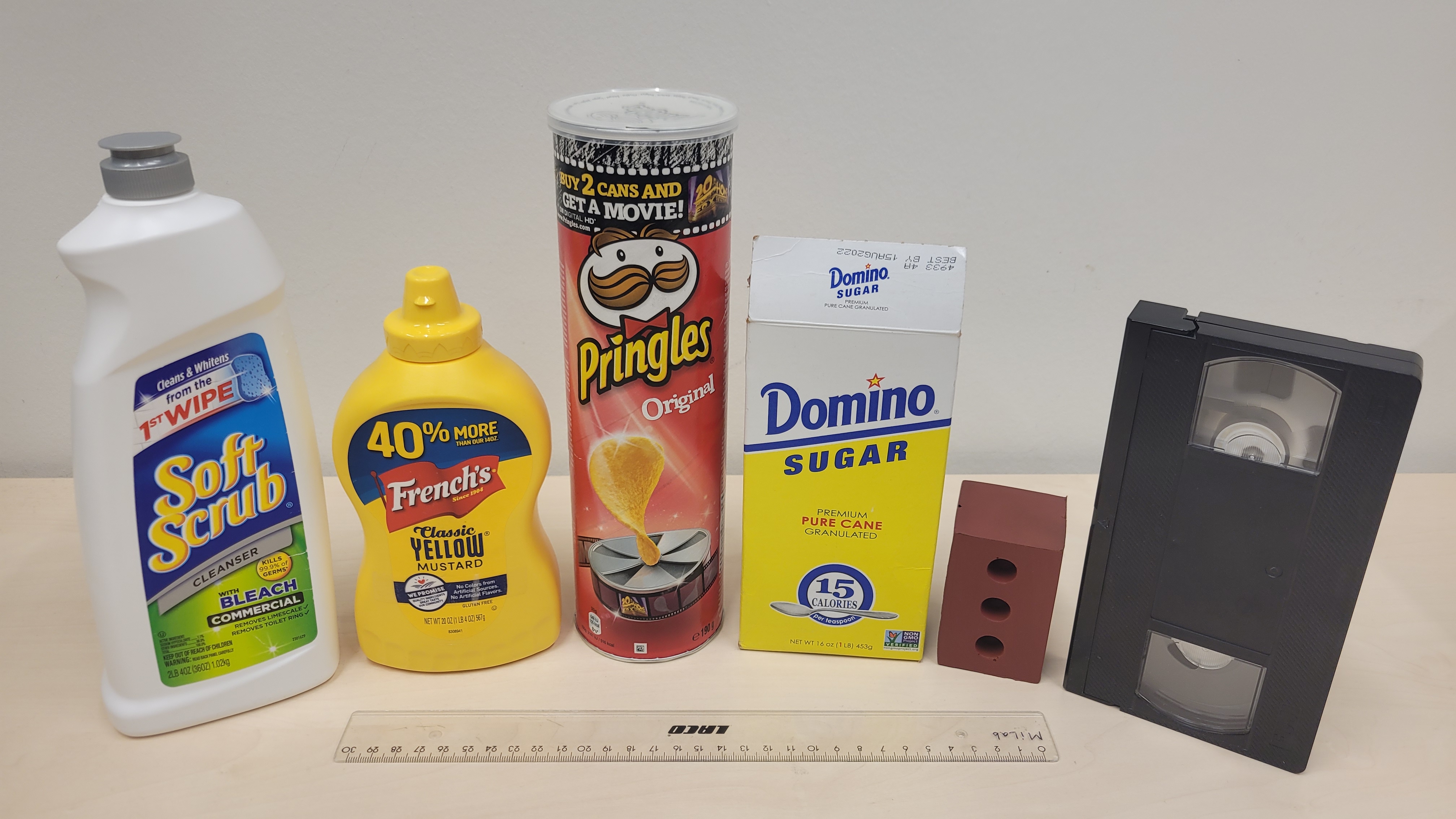}
    \caption{Objects used in the evaluation. From left to right, the objects in the row are bleach cleanser, mustard container, chips can, sugar box, foam brick, and cassette. The object lying in front is the ruler.}
    \label{fig:ycb-objs}
\end{figure}

For the real-world dataset, we took seven differently shaped everyday objects from the YCB object set~\cite{ycbSet} that fit in the robot's gripper: \textit{bleach cleanser}, \textit{foam brick}, \textit{sugar box}, \textit{mustard container}, and \textit{chips can}. We add a cassette and a ruler from our lab equipment to the set to investigate the performance on thin and stiff objects.
All used objects are depicted in figure~\ref{fig:ycb-objs}.

As the recording procedure, we repeated the following protocol for each object:
\begin{enumerate}
    \item Insert the object into the gripper, ensuring the grip is steady enough to avoid slipping too much.
    \item Move the robot gripper to a pre-computed pose $P$ via a simple position controller, such that the end-effector frame (positionally equivalent to the object frame) is located above the sensor's center and $\textbf{R}^W_{O'}$ corresponds to a rotation of $0^{\circ}$ on the sensors z-axis.
    \item Generate a random offset on the xy-plane (parallel to the sensor plane) by sampling both axes individually from $\text{clip}(\mathcal{N}(0, 0.005), -0.2, 0.2)$ and apply it to $P$, to add some translational variation to the dataset. 
    \item Press the object onto the sensor by lowering the gripper on the z-axis to an object-specific offset, resulting in a normal force between $3N$ and $5N$, depending on the contact area.
    \item Save a single sensor reading.
    \item Repeat steps 2-5 30 times, increasing the rotation on the sensor's z-axis by 0.093 rad for each repetition (up to a maximum of 2.8 rad).
\end{enumerate}

This protocol yielded a set of 210 sensor recordings in total for 30 angles between $0^{\circ}$ and $\sim160^{\circ}$ in steps of $\sim5.4^{\circ}$ with varying pressures per object.
Note that the clipping function in step 2 limits the offset to a maximum of four cells on both axes, i.e., a quarter of the sensor's extent, such that the object's center is still within the sensor's bounds.
The sensor values were normalized in [0, 1] as an additional processing step.

\begin{figure}[H]
    \centering
    \includegraphics[width=\linewidth]{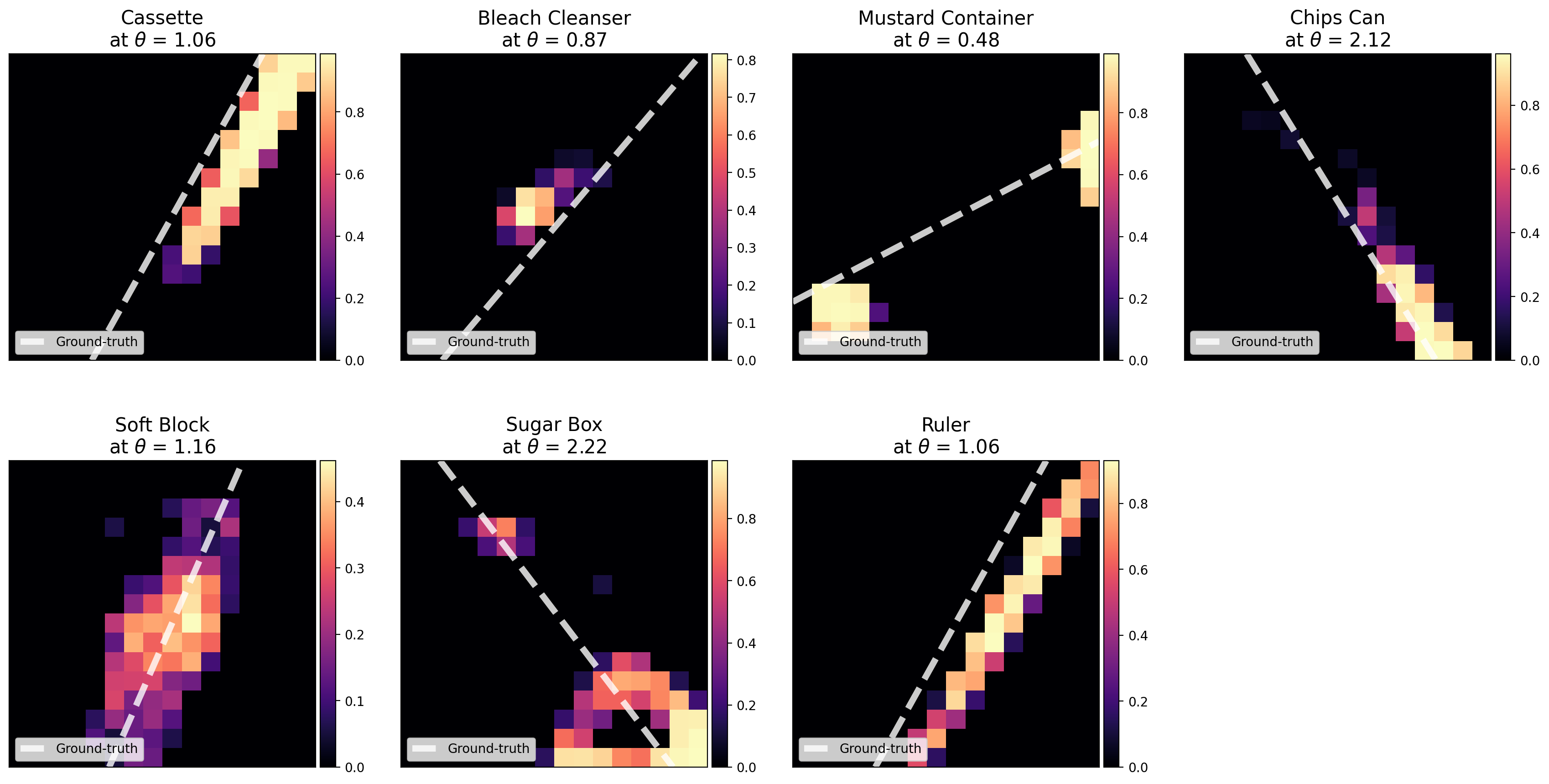}
    \caption{Random samples of real world data collected from the objects depicted in figure~\ref{fig:ycb-objs} along with the respective ground truth orientation as a dotted white line.}
    \label{fig:real_sensordata}
\end{figure}

\begin{figure}[H]
    \centering
    \includegraphics[trim={0 18cm 0 10cm},clip,width=0.8\linewidth]{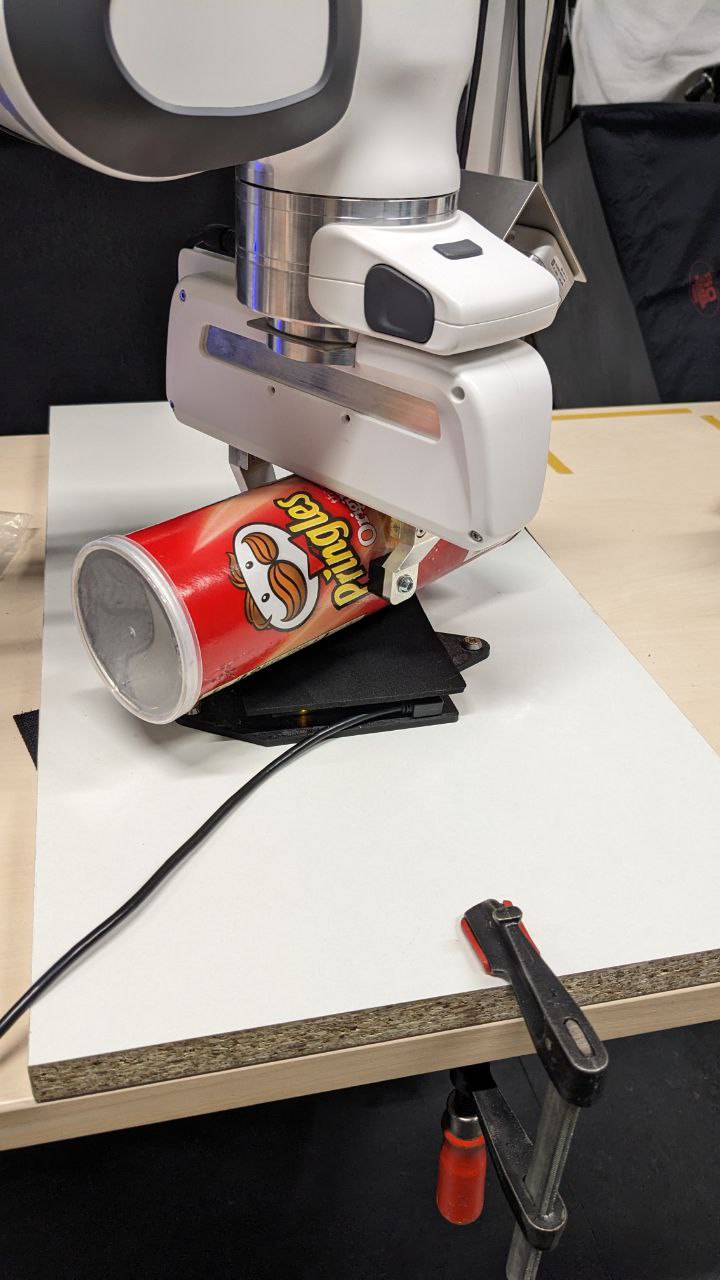}
    \caption{Real world experimental setup. The Panda robot is grasping the \textit{Pringles} YCB object and pressing it against a Myrmex sensor fixed on the table to produce a sensor reading. With the known the rotation of the robots' end-effector, the object rotation can be inferred.}
    \label{fig:panda}
\end{figure}

Some randomly chosen samples of the sensor data recorded for evaluation and the respective ground true object orientation are depicted in figure~\ref{fig:real_sensordata}. Figure~\ref{fig:panda} shows an example of the \textit{Pringles} object being pressed against the sensor.

\subsection{Learning Primitive Orientation in Simulation}

In the simulation, we can neglect the robot control part, as we can directly control the object's positions and orientation through the engine's constraint system. The constraint system employs a parameterized spring-damper model, such that a collision hindering reaching the target pose results in a force proportional to the distance to the target.

\begin{figure*}[bp]
    \centering
    \subfloat[]{\includegraphics[width=0.17\textwidth]{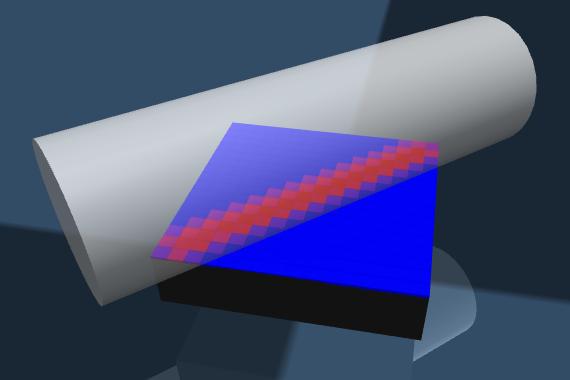}
    \label{fig:show-cyl}}
    \subfloat[]{\includegraphics[width=0.17\textwidth]{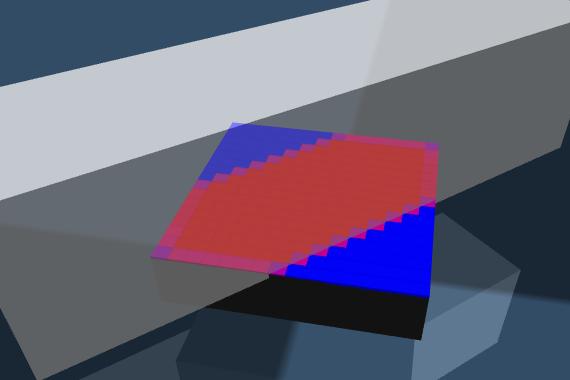}
    \label{fig:show-cub}}
    \subfloat[]{\includegraphics[width=0.17\textwidth]{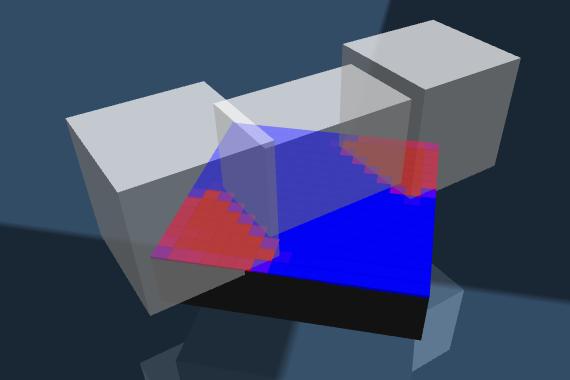}
    \label{fig:show-twocub}}
    \subfloat[]{\includegraphics[width=0.17\textwidth]{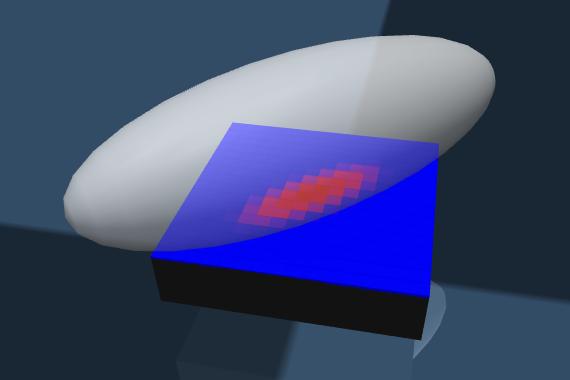}
    \label{fig:show-ell}}
    \subfloat[]{\includegraphics[width=0.17\textwidth]{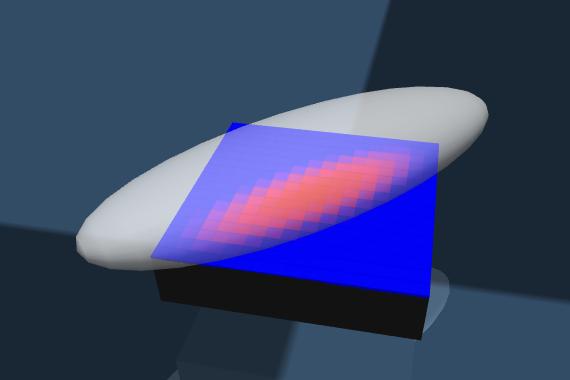}
    \label{fig:show-flat-ell}}
    \caption{Simulated primitives placed on the sensor at an angle of $\pi/4$. From left to right, the objects are cylinder, cuboid, dumbbell, ellipsoid, and flat ellipsoid. On the sensor surface the measured pressure is visualized on a color scale from blue to red, blue indicating no pressure and red high pressure. }
    \label{fig:show_prims}
\end{figure*}

\begin{figure}[H]
    \centering
    \includegraphics[width=\linewidth]{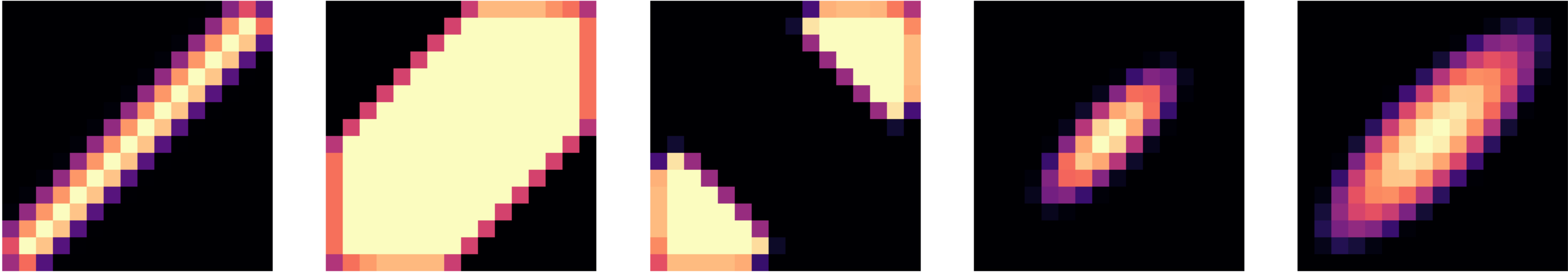}
    \caption{Example sensor data for each primitive object in the training set at an angle of $\pi/4$. From left to right, the objects are cylinder, cuboid, dumbbell, ellipsoid, and flat ellipsoid. The data was recorded for the object placement depicted in figure~\ref{fig:show_prims}.}
    \label{fig:example-data}
\end{figure}

In addition to a cuboid and a cylinder used by~\cite{lachPlacingTouchingEmpirical2023}, our set of primitive training objects consists of a \textit{dumbbell}, an \textit{ellipsoid}, and a \textit{flat ellipsoid}. The objects and an example of their caused sensor responses are displayed in figures \ref{fig:show_prims} and \ref{fig:example-data}, respectively.
The \textit{dumbbell} (figure \ref{fig:show-twocub}) consists of two cuboids connected by a thinner cuboid, such that only the outer cuboids can touch the sensor in the set of poses recorded in this experiment. Therefore, it will always produce either a single or two opposing rectangular clusters in the sensor image.
The \textit{ellipsoid} (figure~\ref{fig:show-ell}) and the \textit{flat ellipsoid} (figure \ref{fig:show-flat-ell}) differ in their axes' radii: while the \textit{ellipsoid} has the same radius on its x and y axes, the \textit{flat ellipsoid} has differing radii resulting in a wider contact surface.
The set of primitive objects was chosen based on the expectation of which patterns can be measured using everyday objects.
For details on the geometries of the simulated objects and their parametrization, we refer to appendix~\ref{app:sim-params}.

For each primitive object, data was recorded for 50 evenly spaced angles in the interval between 0 and $\pi/2$ rad, at a measured reference force of 3, 6, and 9 Newtons, respectively.
Each angle was recorded with 5 random xy-axis translational offsets generated with the same function used in the real setup.

This amounted to 750 samples per object, recorded at an average rate of 25 samples per minute on an Intel Core i7-11700K CPU.
As only a single reading per sample was needed, the recording was run in on-demand mode, computing the sensor values only when requested.

Leveraging sensor symmetry, we duplicated the data set by rotating the data and ground truth angle by $\pi/2$. The resulting data set was then again duplicated by rotating the sensor by $\pi$ but keeping the ground truth angle unchanged due to the invariance to the placing normal's direction ($\Vec{z_p} \equiv -\Vec{z_p}$).
This amounted to a set of synthetic 21.000 samples in total.

Additionally, taking into account the observed pitch described in section~\ref{sec:real-data}, we extended the original by recording every sample with 5 evenly spaced pitch angles between 0 deg and  $\pi/90$ rad ($2^{\circ}$). This resulted in 3.750 samples per object, which were again quadrupled to a total of 105.000 samples.
The former described set is further referred to as the no-pitch, and the latter as the pitch data set.
Both data sets were also normalized in [0, 1], and to better replicate the noisy real-world data, random noise was added to each normalized sensor image sampled from $\mathcal{N}(0.02, 0.05)$.

\subsection{Training}
We trained the neural network architecture on the no-pitch and pitch data sets for 40 epochs with a 20\% validation split, using Adam optimization with a learning rate of $1\mathrm{e}{-3}$. Following the original training protocol from \cite{lachPlacingTouchingEmpirical2023}, the model weights were saved if the running validation loss of the last 10 batches reached a new minimum.

\subsection{Baselines}
As a baseline, we compare our model against the same line-fitting methods used in~\cite{lachPlacingTouchingEmpirical2023}: Principal Component Analysis (PCA)~\cite{FRS1901LIIIOL} and Hough transforms~\cite{10.1145/361237.361242}.

The sensor readings are treated as a bivariate, uni-modal Gaussian for the PCA baseline, and its mean, standard deviations, and covariance matrix C are estimated. The orientation of the object's main axis is obtained by calculating the first principal component of C using PCA.

We transform our sensor readings into a binary image using a noise threshold for the Hough transform baseline. All taxels above the threshold receive a value of 1 all taxels below the threshold receive a value of 0. The hough algorithm finds lines in that image, from which we use the most confident orientation estimate for our object.

\section{Results}
In this section, we report the angular error on the real-world data of the baselines and the neural networks trained in simulation.
We reference the model trained with pitch data as \textit{NN + pitch} and the model trained on the no-pitch set as \textit{NN no-pitch}.
The per-object and average angular errors and standard deviations are listed in table~\ref{tab:results}. Additionally, we also report the medians on the whole data set.
We note that PCA by far shows the best performance on the \textit{soft brick} and also predicts the \textit{ruler} angles with the highest accuracy, followed by Hough and then the NN + pitch model.
Otherwise, the neural networks outperform the baselines on all other objects. While the mean of the NN + pitch model is quite strongly influenced by the much worse performance on the \textit{bleach cleanser} and \textit{Foam Brick}, compared to the remaining objects, its mean performance is very close to the on average better performing no-pitch model. 
Interestingly, the no-pitch model neither performs exceptionally well nor exceptionally poorly on all objects and exhibits the lowest overall standard deviation in predicted errors.
It is also noteworthy to mention that the median angular errors of the neural networks are much lower than the baselines', with the lowest corresponding to about $2.86^{\circ}$.

We also briefly compare our data with the original results of Lach et al. on predicting previously unseen everyday objects, also stemming from the YCB set.
They reach a mean angular error of 0.09 rad with a 0.07 rad standard deviation, which is very similar to our results even though we use half the sensor input for an almost identical task.
This leads us to deem our sim-to-real validation a success, noticeably without any domain randomization or augmentation methods that exceed gaussian noise.

\begin{table*}[!t]
\caption{Experimental results for seven previously unknown household objects (see figure \ref{fig:ycb-objs}). Angular errors are reported in radians. The Hough algorithm cannot predict any orientation for the bleach cleanser object. Hence, no value is displayed.}
\label{tab:results}
\centering
\begin{tabular}{|l||p{2cm}|p{1cm}|p{1cm}|p{1cm}|p{1cm}|p{1cm}|p{1cm}||p{1cm}|p{1cm}|}
\hline
Model & Bleach Cleanser& Foam Brick& Sugar Box& Mustard Container& Chips Can& Cassette & Ruler&average &median\\
\hline
NN + Pitch& \textbf{0.07±0.0}& 0.2±0.02& 0.16±0.01& \textbf{0.1±0.0}& \textbf{0.05±0.0}& \textbf{0.06±0.0}& 0.07±0.01&0.1±0.09&\textbf{0.05}\\\hline
\hline
 NN no-pitch& 0.11±0.0& 0.14±0.01& \textbf{0.09±0.0}& 0.1±0.01& 0.09±0.0& 0.1±0.01& 0.11±0.01& \textbf{0.1±0.08}&0.07\\\hline
\hline
 PCA& 0.16±0.15& \textbf{0.08±0.05}& 0.13±0.15& 0.14±0.22& 0.05±0.04& 0.19±0.28& \textbf{0.04±0.02}& 0.11±0.13& 0.13\\\hline
 Hough& -& 0.13±0.12& 0.25±0.26& 0.22±0.0& 0.08±0.1& 0.09±0.16& 0.06±0.06& 0.14±0.12& 0.11 \\\hline
\end{tabular}
\end{table*}

\section{Discussion}

When looking at the sensor data recorded for the objects, the data of the chips can and ruler seem the easiest to estimate the object angle from. Although the chips can is the wider object, both leave an imprint resembling a slim but well-recognizable line in the sensor image. Hence, the line-fitting baseline models perform well on them.

Because objects in the gripper had some leeway and the containers were empty, they tended to move when pushed on the sensor, resulting in slight rotations on the axis orthogonal to the gripper's and object's z-axis (pitch). 
This sometimes caused situations where the object was only \textit{visible} in the 5-cells outer boundary of the sensor on one side of the array.
Compared to the no-pitch model, the model trained on the pitch data performs better on objects we most expected to see pitched samples, namely the ruler and the cassette.

In the gripper, the foam brick was occasionally strongly deformed, as it is the softest evaluation object. Various contact surface patterns resulted, some of which remotely resemble an ellipsoid bent along its primary axis. Thus, we believe its challenging sensor patterns are the reason why both networks display their worst performance on this object. It would be interesting to see if adding varying object softness to the synthetic data would result in a decreased error for the \textit{foam block}.

The contact surface with the myrmex sensor of the sugar box and the mustard container consists of two non-connected segments that are both on opposing edges of the sensor. From the shapes of the object, we expected such patterns and, therefore, introduced the dumbbell primitive into the training set. However, sometimes, one of the segments is barely or not visible on the sensor at all, most likely due to the object's pitch. 
The sugar box also produced a similar shape because it has stiff edges but less support on the sides and, therefore, is deformed by the gripper in such a way that the middle part is dented inwards. 
We can not explain why the no-pitch model reaches a moderate performance of 0.9 rad angular error while all other estimators seem to struggle with the data from this object.

We cautiously compared our data with the original results of Lach et al. \cite{lachPlacingTouchingEmpirical2023} on predicting previously unseen everyday objects, also stemming from the YCB set.

\subsection{Generalizability to other Sensors}

Our sensor simulation applies to different types of existing tactile sensors besides the piezoresistive sensor used in our experiment.  
For example, optical sensors could be emulated by rendering the contact surface corresponding to the deformed sensor surface. 

\begin{figure}[tpb]
    \centering
    \subfloat[]{\includegraphics[width=0.5\linewidth]{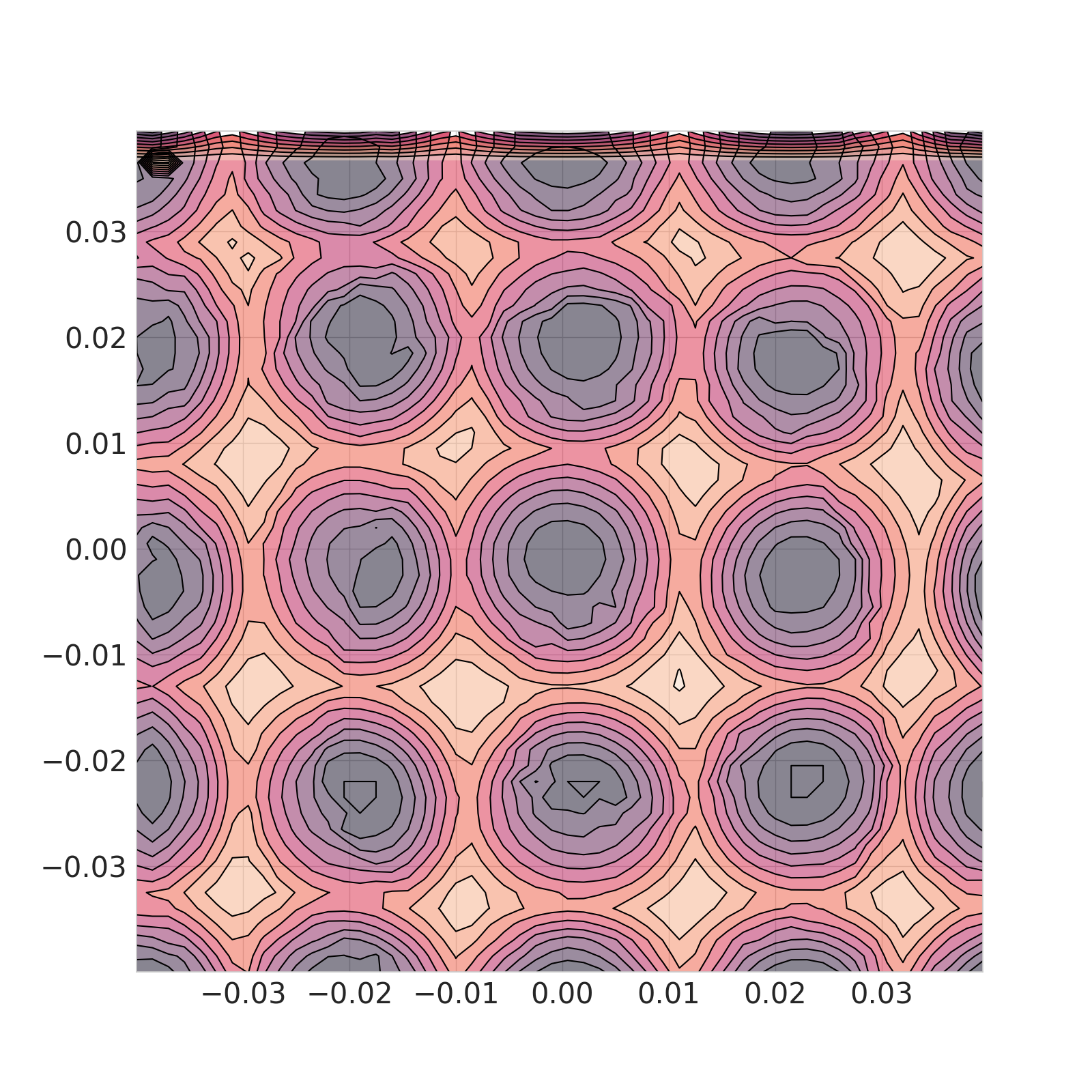}}
    \subfloat[]{\includegraphics[width=0.5\linewidth]{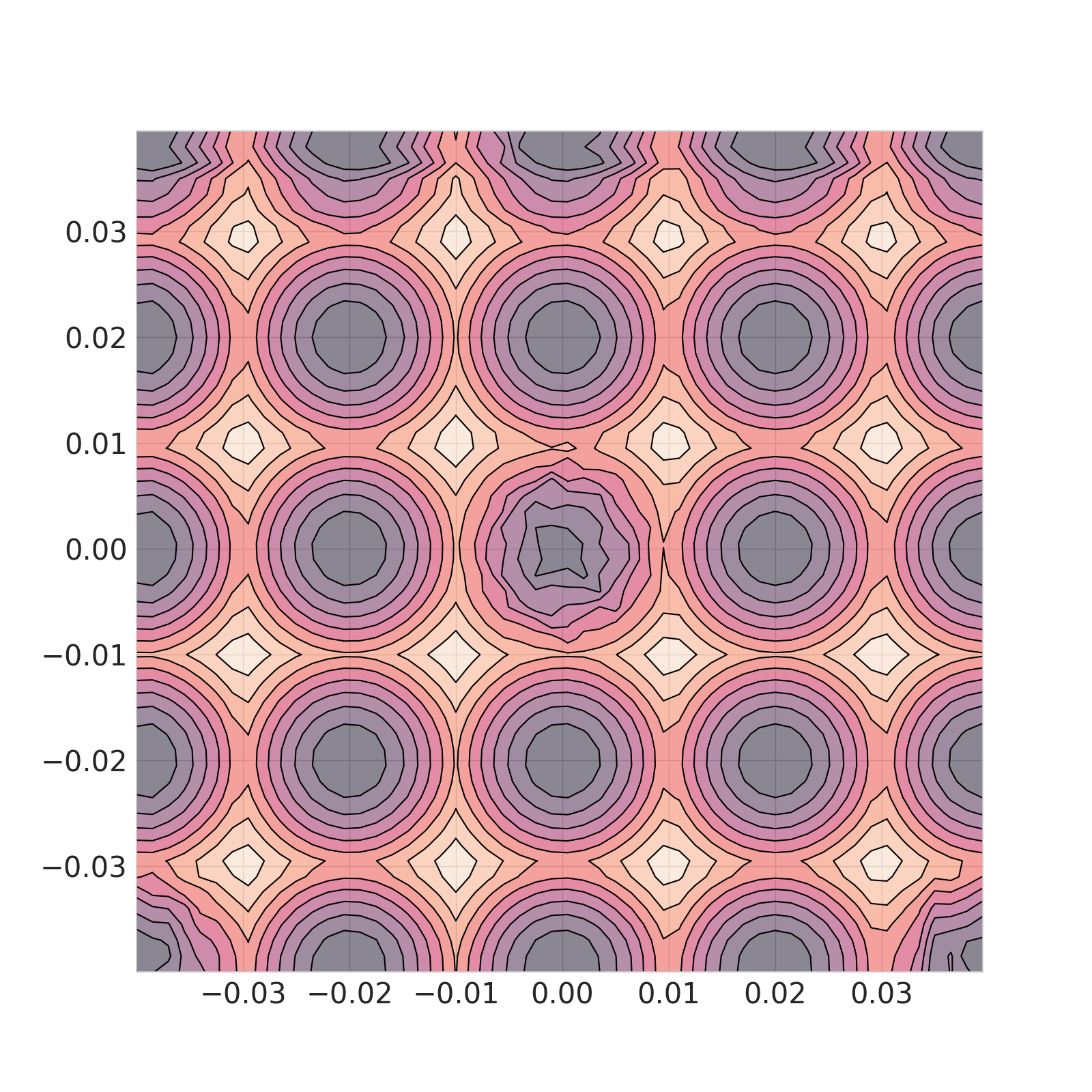}}
    \caption{Sensor data from a 5x5 barometric sensor array~\cite{koiva2019baro5x5} recorded in a real (a) and simulated (b) setup. A cylindrical probe tip was pressed on the sensor surface with a force of 0.35~Newton along a grid that covers the full sensor surface. The pictures show the sensor responses for all sensor cells overlayed for all measuring positions. Darker colors represent more force being measured by the tactile sensor.}
    \label{fig:baro_sensor}
\end{figure}

In a preliminary study~\cite{leins2023more}, we already did a qualitative evaluation of a 5x5 barometric tactile sensor array~\cite{koiva2019baro5x5}. In that study, we only measured and compared real and synthetic sensor data on one of the sensors' main axes. We now extended this evaluation and recorded data across the full sensor surface. The sensor data was recorded using a 3-axis stepper motor setup carrying a cylindrical probe tip ($\diameter$ 2~mm) attached to a spring scale. The probe tip traversed the sensor surface with a step size of 0.15~mm. The tip was lowered for each position until a reference force of 0.35~Newton was measured. Replacing the 3-axis stepper motor with constraints, the same data was recorded for a simulated sensor.
A qualitative comparison between the synthetic and real sensor data is depicted in figure~\ref{fig:baro_sensor}.

In another preliminary study~\cite{patzeltCurvedTactileSensor2023}, we showcased a piezoresistive fingertip sensor simulation for the Shadow Dexterous Hand~\cite{Koiva2013AHS}. This highlights, that our sensor simulation is not restricted to flat sensor surfaces but can be readily applied to arbitrary sensor shapes. We achieve this by moving from 2D taxel coordinates on the sensor surface to 3D taxel locations and normals on the sensor mesh. Suitable sample points for the raycasting are generated using Constrained Poisson-disk sampling as described in section~\ref{sec:deriving_pressure}.

\section{Conclusion and Future Work}
In summary, our study introduces a tactile sensor simulation within our MuJoCo-based simulation framework, employing hydroelastic contact surfaces. Through the aggregation of pressure values in the receptive region of a sensor, our model yields improved estimates of force distribution compared to traditional point contact models. Our simulation allows for (faster than) real-time speed for a common tactile sensor array, thus offering a pragmatic alternative to slow but more precise Finite Element methods. 
Our simulation's adaptability through parametrization is a crucial feature, providing a customizable trade-off between computational efficiency and accuracy. 

Validation through a sim-to-real transfer experiment demonstrates the utility of our sensor simulation: A neural network trained exclusively on synthetic data accurately predicts object states on a physical sensor, showcasing the model's efficacy in capturing tactile interactions for real-world applications.

The main limitation of our implementation lies in its inability to scale with bigger sensor configurations or high sampling rates because of the quadratically growing computational effort. To mitigate this, we plan to move the raycasting pipeline onto a GPU to allow a much faster parallelization factor on a much higher core count.

In the future, we also plan to perform more validation experiments with other sensor modalities and include leveraging our method's capability of simulating soft-to-soft collisions to further narrow the sim-to-real gap in soft tactile interaction.

\section*{Acknowledgments}
We would like to thank Luca Lach for providing insights into their previous work and helping set up the experimental pipeline.
This work was supported by the BMBF project Sim4Dexterity.

{\appendices
\section{Simulation Parameters}
\label{app:sim-params}
\paragraph{Sensor} The sensor is simulated by a 8x8x1.8~cm box with 16 by 16 taxels evenly distributed on its upward facing surface. As sensor parameters we chose a resolution of 20 samples per axis, a hydroelastic modulus of $3*10^4$~N/m$^2$, a dissipation of 20~s/m, and static and dynamic friction coefficients of 0.3

\paragraph{Training Objects} The cuboid is a 5x5x19~cm box with a resolution hint of 0.05. The cylinder has a length of 15~cm, a radius of 2.25~cm and is modeled with a resolution hint of 0.01. The ellipsoids are scaled spheres with radii of 2.5x2.5x7.5~cm and 2.5x1.5x7.5~cm. Both ellipsoids are modeled with a resolution hint of 0.02. The dumbbell consists of 3 boxes. The outer boxes are 4x4x4~cm respectively, the handle connecting them is 3x3x5~cm. All boxes forming the dumbbell have a resolution hint of 0.05. 
As the objects are held into place by constraints they are all modeled as friction-less. All training objects are modeled as rigid, so the hydroelastic modulus can be omitted and the dissipation is set to 0.

\bibliographystyle{IEEEtran}
\bibliography{Tactile}

\begin{thebibliography}{10}
\providecommand{\url}[1]{#1}
\csname url@samestyle\endcsname
\providecommand{\newblock}{\relax}
\providecommand{\bibinfo}[2]{#2}
\providecommand{\BIBentrySTDinterwordspacing}{\spaceskip=0pt\relax}
\providecommand{\BIBentryALTinterwordstretchfactor}{4}
\providecommand{\BIBentryALTinterwordspacing}{\spaceskip=\fontdimen2\font plus
\BIBentryALTinterwordstretchfactor\fontdimen3\font minus
  \fontdimen4\font\relax}
\providecommand{\BIBforeignlanguage}[2]{{%
\expandafter\ifx\csname l@#1\endcsname\relax
\typeout{** WARNING: IEEEtran.bst: No hyphenation pattern has been}%
\typeout{** loaded for the language `#1'. Using the pattern for}%
\typeout{** the default language instead.}%
\else
\language=\csname l@#1\endcsname
\fi
#2}}
\providecommand{\BIBdecl}{\relax}
\BIBdecl

\bibitem{mandilTactileSensingTechnologiesTrends2023}
W.~Mandil, V.~Rajendran, K.~Nazari, and A.~Ghalamzan-Esfahani,
  ``Tactile-{{Sensing Technologies}}: {{Trends}}, {{Challenges}} and
  {{Outlook}} in {{Agri-Food Manipulation}},'' vol.~23, no.~17, p. 7362.

\bibitem{wangHSVTacHighSpeedVisionBased2023}
X.~Wang, Y.~Yang, Z.~Zhou, G.~Xiang, and H.~Liu, ``{{HSVTac}}: {{A High-Speed
  Vision-Based Tactile Sensor}} for {{Exploring Fingertip Tactile
  Sensitivity}},'' vol.~23, no.~19, pp. 23\,431--23\,439.

\bibitem{piacenzaSensorizedMulticurvedRobot2020}
P.~Piacenza, K.~Behrman, B.~Schifferer, I.~Kymissis, and M.~Ciocarlie, ``A
  {{Sensorized Multicurved Robot Finger With Data-Driven Touch Sensing}} via
  {{Overlapping Light Signals}},'' vol.~25, no.~5, pp. 2416--2427.

\bibitem{gomesGelTipFingershapedOptical2020}
D.~F. Gomes, Z.~Lin, and S.~Luo, ``{{GelTip}}: {{A Finger-shaped Optical
  Tactile Sensor}} for {{Robotic Manipulation}},'' in \emph{2020 {{IEEE}}/{{RSJ
  International Conference}} on {{Intelligent Robots}} and {{Systems}}
  ({{IROS}})}, pp. 9903--9909.

\bibitem{lambetaDIGITNovelDesign2020}
M.~Lambeta, P.-W. Chou, S.~Tian, B.~Yang, B.~Maloon, V.~R. Most, D.~Stroud,
  R.~Santos, A.~Byagowi, G.~Kammerer, D.~Jayaraman, and R.~Calandra,
  ``{{DIGIT}}: {{A Novel Design}} for a {{Low-Cost Compact High-Resolution
  Tactile Sensor With Application}} to {{In-Hand Manipulation}},'' vol.~5,
  no.~3, pp. 3838--3845.

\bibitem{padmanabhaOmniTactMultiDirectionalHighResolution2020}
A.~Padmanabha, F.~Ebert, S.~Tian, R.~Calandra, C.~Finn, and S.~Levine,
  ``{{OmniTact}}: {{A Multi-Directional High-Resolution Touch Sensor}},'' in
  \emph{2020 {{IEEE International Conference}} on {{Robotics}} and
  {{Automation}} ({{ICRA}})}.\hskip 1em plus 0.5em minus 0.4em\relax {IEEE},
  pp. 618--624.

\bibitem{zhuRecentAdvancesResistive2022}
Y.~Zhu, Y.~Liu, Y.~Sun, Y.~Zhang, and G.~Ding, ``Recent {{Advances}} in
  {{Resistive Sensor Technology}} for {{Tactile Perception}}: {{A Review}},''
  vol.~22, no.~16, pp. 15\,635--15\,649.

\bibitem{zhaoSimtoRealTransferDeep2020}
W.~Zhao, J.~P. Queralta, and T.~Westerlund, ``Sim-to-{{Real Transfer}} in
  {{Deep Reinforcement Learning}} for {{Robotics}}: A {{Survey}},'' in
  \emph{2020 {{IEEE Symposium Series}} on {{Computational Intelligence}}
  ({{SSCI}})}, pp. 737--744.

\bibitem{gomesGelSightSimulationSim2Real2019}
D.~F. Gomes, A.~Wilson, and S.~Luo, ``{{GelSight Simulation}} for {{Sim2Real
  Learning}}.''

\bibitem{dingSimtoRealTransferOptical2020}
Z.~Ding, N.~F. Lepora, and E.~Johns, ``Sim-to-{{Real Transfer}} for {{Optical
  Tactile Sensing}},'' in \emph{2020 {{IEEE International Conference}} on
  {{Robotics}} and {{Automation}} ({{ICRA}})}, pp. 1639--1645.

\bibitem{sferrazzaLearningSenseTouch2020}
C.~Sferrazza, T.~Bi, and R.~D’Andrea, ``Learning the sense of touch in
  simulation: A sim-to-real strategy for vision-based tactile sensing,'' in
  \emph{2020 {{IEEE}}/{{RSJ International Conference}} on {{Intelligent
  Robots}} and {{Systems}} ({{IROS}})}, pp. 4389--4396.

\bibitem{churchTactileSimtoRealPolicy2022}
A.~Church, J.~Lloyd, R.~Hadsell, and N.~F. Lepora, ``Tactile {{Sim-to-Real
  Policy Transfer}} via {{Real-to-Sim Image Translation}},'' in
  \emph{Proceedings of the 5th {{Conference}} on {{Robot Learning}}}.\hskip 1em
  plus 0.5em minus 0.4em\relax {PMLR}, pp. 1645--1654.

\bibitem{siTaximExampleBasedSimulation2022}
Z.~Si and W.~Yuan, ``Taxim: {{An Example-Based Simulation Model}} for
  {{GelSight Tactile Sensors}},'' vol.~7, no.~2, pp. 2361--2368.

\bibitem{wangTACTOFastFlexible2022}
S.~Wang, M.~Lambeta, P.-W. Chou, and R.~Calandra, ``{{TACTO}}: {{A Fast}},
  {{Flexible}}, and {{Open-Source Simulator}} for {{High-Resolution
  Vision-Based Tactile Sensors}},'' vol.~7, no.~2, pp. 3930--3937.

\bibitem{xuEfficientTactileSimulation2022}
J.~Xu, S.~Kim, T.~Chen, A.~R. Garcia, P.~Agrawal, W.~Matusik, and S.~Sueda,
  ``Efficient {{Tactile Simulation}} with {{Differentiability}} for {{Robotic
  Manipulation}}.''

\bibitem{chenTacchiPluggableLow2023}
Z.~Chen, S.~Zhang, S.~Luo, F.~Sun, and B.~Fang, ``Tacchi: {{A Pluggable}} and
  {{Low Computational Cost Elastomer Deformation Simulator}} for {{Optical
  Tactile Sensors}},'' vol.~8, no.~3, pp. 1239--1246.

\bibitem{chenGeneralPurposeSim2RealProtocol2024}
W.~Chen, J.~Xu, F.~Xiang, X.~Yuan, H.~Su, and R.~Chen, ``General-{{Purpose
  Sim2Real Protocol}} for {{Learning Contact-Rich Manipulation With
  Marker-Based Visuotactile Sensors}},'' pp. 1--18.

\bibitem{narangInterpretingPredictingTactile2020}
Y.~S. Narang, K.~Van~Wyk, A.~Mousavian, and D.~Fox. Interpreting and
  {{Predicting Tactile Signals}} via a {{Physics-Based}} and {{Data-Driven
  Framework}}.

\bibitem{narangSimtoRealRoboticTactile2021}
Y.~Narang, B.~Sundaralingam, M.~Macklin, A.~Mousavian, and D.~Fox,
  ``Sim-to-{{Real}} for {{Robotic Tactile Sensing}} via {{Physics-Based
  Simulation}} and {{Learned Latent Projections}},'' in \emph{2021 {{IEEE
  International Conference}} on {{Robotics}} and {{Automation}} ({{ICRA}})},
  pp. 6444--6451.

\bibitem{kappassovSimulationTactileSensing2020}
Z.~Kappassov, J.-A. Corrales-Ramon, and V.~Perdereau, ``Simulation of {{Tactile
  Sensing Arrays}} for {{Physical Interaction Tasks}},'' in \emph{2020
  {{IEEE}}/{{ASME International Conference}} on {{Advanced Intelligent
  Mechatronics}} ({{AIM}})}, pp. 196--201.

\bibitem{elandtPressureFieldModel2019}
R.~Elandt, E.~Drumwright, M.~Sherman, and A.~Ruina, ``A pressure field model
  for fast, robust approximation of net contact force and moment between
  nominally rigid objects,'' in \emph{2019 {{IEEE}}/{{RSJ International
  Conference}} on {{Intelligent Robots}} and {{Systems}} ({{IROS}})}, pp.
  8238--8245.

\bibitem{todorov2012mujoco}
E.~Todorov, T.~Erez, and Y.~Tassa, ``Mujoco: A physics engine for model-based
  control,'' in \emph{2012 IEEE/RSJ International Conference on Intelligent
  Robots and Systems}.\hskip 1em plus 0.5em minus 0.4em\relax IEEE, 2012, pp.
  5026--5033.

\bibitem{wettelsBiomimeticTactileSensor2008}
N.~Wettels, V.~J. Santos, R.~S. Johansson, and G.~E. Loeb, ``Biomimetic
  {{Tactile Sensor Array}},'' vol.~22, no.~8, pp. 829--849.

\bibitem{ward-cherrierTacTipFamilySoft2018}
B.~Ward-Cherrier, N.~Pestell, L.~Cramphorn, B.~Winstone, M.~E. Giannaccini,
  J.~Rossiter, and N.~F. Lepora, ``The {{TacTip Family}}: {{Soft Optical
  Tactile Sensors}} with {{3D-Printed Biomimetic Morphologies}},'' vol.~5,
  no.~2, pp. 216--227.

\bibitem{wangElasticTactileSimulation2021}
Y.~Wang, W.~Huang, B.~Fang, F.~Sun, and C.~Li, ``Elastic {{Tactile Simulation
  Towards Tactile-Visual Perception}},'' in \emph{Proceedings of the 29th {{ACM
  International Conference}} on {{Multimedia}}}, ser. {{MM}} '21.\hskip 1em
  plus 0.5em minus 0.4em\relax {Association for Computing Machinery}, pp.
  2690--2698.

\bibitem{masterjohnVelocityLevelApproximation2022}
J.~Masterjohn, D.~Guoy, J.~Shepherd, and A.~Castro, ``Velocity {{Level
  Approximation}} of {{Pressure Field Contact Patches}},'' vol.~7, no.~4, pp.
  11\,593--11\,600.

\bibitem{moller2005fast}
T.~M{\"o}ller and B.~Trumbore, ``Fast, minimum storage ray/triangle
  intersection,'' in \emph{SIGGRAPH 2005 Courses}.\hskip 1em plus 0.5em minus
  0.4em\relax ACM, 2005.

\bibitem{wald2007fast}
I.~Wald, ``On fast construction of sah-based bounding volume hierarchies,'' in
  \emph{2007 IEEE Symposium on Interactive Ray Tracing}.\hskip 1em plus 0.5em
  minus 0.4em\relax IEEE, 2007, pp. 33--40.

\bibitem{sampling}
M.~Corsini, P.~Cignoni, and R.~Scopigno, ``Efficient and flexible sampling with
  blue noise properties of triangular meshes,'' \emph{IEEE Transactions on
  Visualization and Computer Graphics}, vol.~18, no.~6, 2012.

\bibitem{leins2023more}
D.~Leins, F.~Patzelt, and R.~Haschke, ``More accurate tactile sensor simulation
  with hydroelastic contacts in mujoco,'' in \emph{IROS 2023 Workshop on
  Leveraging Models for Contact-Rich Manipulation}, 2023.

\bibitem{lachPlacingTouchingEmpirical2023}
L.~Lach, N.~Funk, R.~Haschke, S.~Lemaignan, H.~J. Ritter, J.~Peters, and
  G.~Chalvatzaki, ``Placing by touching: An empirical study on the importance
  of tactile sensing for precise object placing,'' in \emph{2023 IEEE/RSJ
  International Conference on Intelligent Robots and Systems (IROS)}.\hskip 1em
  plus 0.5em minus 0.4em\relax IEEE, 2023, pp. 8964--8971.

\bibitem{SchurmannMyrmex}
C.~Schurmann, R.~K~oiva, R.~Haschke, and H.~Ritter, ``A modular high-speed
  tactile sensor for human manipulation research,'' in \emph{2011 IEEE World
  Haptics Conference}, 2011, pp. 339--344.

\bibitem{ycbSet}
B.~Calli, A.~Singh, A.~Walsman, S.~Srinivasa, P.~Abbeel, and A.~M. Dollar,
  ``The ycb object and model set: Towards common benchmarks for manipulation
  research,'' in \emph{2015 International Conference on Advanced Robotics
  (ICAR)}, 2015, pp. 510--517.

\bibitem{FRS1901LIIIOL}
K.~P. F.R.S., ``Liii. on lines and planes of closest fit to systems of points
  in space,'' \emph{Philosophical Magazine Series 1}, vol.~2, pp. 559--572,
  1901.

\bibitem{10.1145/361237.361242}
R.~O. Duda and P.~E. Hart, ``Use of the hough transformation to detect lines
  and curves in pictures,'' \emph{Commun. ACM}, vol.~15, no.~1, p. 11–15, jan
  1972.

\bibitem{koiva2019baro5x5}
R.~K{\~o}iva, T.~Schwank, R.~Haschke, and H.~Ritter, ``Towards high-density
  barometer-based tactile sensor arrays,'' in \emph{Workshop on New Advances in
  Tactile Sensation, Perception, and Learning in Robotics: Emerging Materials
  and Technologies for Manipulation (RoboTac 2019)}, 2019.

\bibitem{patzeltCurvedTactileSensor2023}
F.~Patzelt, D.~Leins, and R.~Haschke, ``Curved {{Tactile Sensor Simulation}}
  with {{Hydroelastic Contacts}} in {{MuJoCo}},'' in \emph{NeurIPS 2023
  Workshop on Touch Processing: a new Sensing Modality for AI}.

\bibitem{Koiva2013AHS}
\BIBentryALTinterwordspacing
R.~K{\~o}iva, M.~Zenker, C.~Sch{\"u}rmann, R.~Haschke, and H.~J. Ritter, ``A
  highly sensitive 3d-shaped tactile sensor,'' \emph{IEEE/ASME Int. Conf. on
  Advanced Intelligent Mechatronics}, pp. 1084--1089, 2013. [Online].
  Available: \url{https://api.semanticscholar.org/CorpusID:9450523}
\BIBentrySTDinterwordspacing

\end{thebibliography}

\end{document}